\documentclass{article}
\usepackage[preprint]{neurips_2018}

\usepackage{epsfig}
\usepackage{graphicx, subfigure}
\usepackage{float}
\usepackage{cite}
\usepackage{natbib}
\usepackage{amsmath}
\usepackage{amssymb}
\usepackage{bm}
\usepackage{enumerate}
\usepackage{epstopdf}
\usepackage{url}
\usepackage{algorithm, algorithmic}
\usepackage{underscore} 
\usepackage[utf8]{inputenc} 
\usepackage[T1]{fontenc}    
\usepackage{hyperref}       
\usepackage{url}            
\usepackage{booktabs}       
\usepackage{amsfonts}       
\usepackage{nicefrac}       
\usepackage{microtype}      

\title
{Learning Nonlinear State Space Models with Hamiltonian Sequential Monte Carlo Sampler}

\author{%
  Duo Xu\thanks{} \\
  Department of Electrical and Computer Engineering\\
  Georgian Institute of Technology\\
  Atlanta, GA 30332 \\
}

\begin{document}

\maketitle

\begin{abstract}
State space models (SSM) have been widely applied for the analysis and visualization of large sequential datasets. Sequential Monte Carlo (SMC) is a very popular particle-based method to sample latent states from intractable posteriors. However, SSM is significantly influenced by the choice of the proposal. Recently Hamiltonian Monte Carlo (HMC) sampling has shown success in many practical problems. In this paper, we propose an SMC augmented by HMC (HSMC) for inference and model learning of nonlinear SSM, which can exempt us from learning proposals and reduce the model complexity significantly. Based on the measure preserving property of HMC, the particles directly generated by transition function can approximate the posterior of latent states arbitrarily well. In order to better adapt to the local geometry of latent space, the HMC is conducted on Riemannian manifold defined by a positive definite metric $\bm{M}(\bm{x})$. In addition, we show that the proposed HSMC method can improve SSMs realized by both Gaussian Processes (GP) and Neural Network (NN).
\end{abstract}

\section{Introduction}
System identification \citep{ljung1998system, ljung2010perspectives} is a fundamental ingredient of many problems, such as model-predictive control \citep{camacho2013model} and model-based reinforcement learning \citep{deisenroth2011pilco, berkenkamp2017safe}, which is to learn system dynamics from practical data. State-space model (SSM) \citep{billings2013nonlinear} is the most popular method to represent the system with input $\bm{u}_t$ and output $\bm{y}_t$ as functions of a latent Markovian state $\bm{x}_t$. Specifically, linear and non-linear Gaussian state space models (GSSM) are most widely used in practical applications from robotic planning to neural signal processing. However, despite significant effort in research community over past decades, efficient learning method for non-linear GSSM is still lacking.

The sequential Monte Carlo (SMC) \citep{gordon1993novel} is a classical method to infer latent state in SSM, which uses weighted particles to approximate the intractable posterior of the latent states. The proposal distribution, from which particles are sampled, has significant influence on the approximation performance. In order to perform model learning and proposal adaptation at the same time, recent work \citep{maddison2017filtering, le2017auto, naesseth2017variational} combines variational auto-encoder (VAE) \citep{kingma2013auto, rezende2014stochastic} with importance weighted auto-encoder (IWAE) \citep{burda2015importance}, and uses SMC as the estimator for marginal observation likelihood. However, since in nonlinear GSSM the emission and transition frameworks are realized by deep neural networks, the real posterior of latent states is intractable to sample, and the proposal distributions in previous work are always assumed to be Gaussian, different from the true posterior. Although the approximated log-likelihood is unbiased asymptotically \citep{maddison2017filtering}, it is not close to the real log-likelihood without large number of particles. 

In order to improve the inference performance of SMC with low sampling and model complexity, here we propose an SMC sampler augmented by Hamiltonian dynamics (HSMC). Different from previous SMC methods, we don't need proposals here to generate particles of latent states. Using transition function as initial distribution, we use HMC to sample particles to approximate the posterior. And since latent space is time-dependent in nonlinear SSM, we modified the Riemannian Manifold HMC (RMHMC) \citep{girolami2011riemann}, whose mass matrix is generated by an MLP with latent states as inputs. The state dynamics in SSM is usually realized by Gaussian Processes (GP) and neural network. GP is data-efficient and has much less parameters than neural network, but its expressive capability is not as good as neural network, especially in high-dimensional problems. Here we show that the proposed method can improve the learning performance of GP to be comparable as neural network. 

\section{Preliminary}
Denote $\bm{u}_t, \bm{x}_t$ and $\bm{y}_t$ as input, latent and output variables at time $t$ respectively.
\subsection{Gaussian State Space Model}
Gaussian state space model (GSSM) is the most popular method to model the dynamics of complex sequential data in the latent space \citep{raiko2009variational}. The inference and learning of GSSM are both considered in this work. The model is defined as
\begin{eqnarray}
\bm{x}_t&\sim&\mathcal{N}(\mu_{\theta}(\bm{x}_{t-1}, \bm{u}_t), \sigma^2_{\theta}(\bm{x}_{t-1}, \bm{u}_t)) \hspace{25pt}\text{Transition}\nonumber \\
\bm{y}_t&\sim&\Pi(f_{\theta}(\bm{x}_{t-1}, \bm{u}_t))  \hspace{90pt}\text{Emission}\label{model}
\end{eqnarray}
where $\Pi$ is the output distribution parametrized by function $f$, and latent variables distribute as multivariate Gaussian conditioned on previous latent variables and input variables. The GSSM defined above includes both linear and nonlinear GSSM. When functions $f$ and $\mu$ are linear, the model can be learned by extended Kalman filter \citep{wan1997dual} and expectation maximization \citep{ghahramani1999learning}. However, in most practical problems, the dynamic and emission functions are nonlinear. In this paper we propose an efficient method to deal with nonlinearties.

\subsection{Variational Sequential Monte Carlo}
SMC performs particle-based approximate inference on a sequence of target distributions, which is the extension of importance sampling to sequential data. In the context of SSM, the target distributions are the posterior of latent variables, i.e., $\{p_{\theta}(\bm{x}_{1:t}|\bm{y}_{1:t})\}_{t=1}^T$. The generative model includes initial distribution of latent variables $p_{\theta}(\bm{x}_0)$, transition distribution $p_{\theta}(\bm{x}_{t}|\bm{x}_{t-1}, \bm{u}_t)$, and emission distribution $p_{\theta}(\bm{y}_t|\bm{x}_t)$. The proposal distribution, defined as $q_{\phi}(\bm{x}_t|\bm{x}_{t-1}, \bm{y}_t, \bm{u}_t)$, is the approximate inference on target distributions. Then the generative and inference models are factorized as
\begin{eqnarray}
p_{\theta}(\bm{x}_{\le T}, \bm{y}_{\le T}|\bm{u}_{\le T})&=&p_{\theta}(\bm{x}_0)\prod_{t=1}^T p_{\theta}(\bm{x}_t|\bm{x}_{t-1},\bm{u}_t)p_{\theta}(\bm{y}_t|\bm{x}_t) \nonumber \\
q_{\theta}(\bm{x}_{\le T}|\bm{y}_{\le T}, \bm{u}_{\le T})&=&p_{\theta}(\bm{x}_0)\prod_{t=1}^T q_{\theta}(\bm{x}_t|\bm{x}_{t-1},\bm{y}_t, \bm{u}_t) \nonumber
\end{eqnarray}
Recently, a new ELBO objective has been introduced \citep{le2017auto, maddison2017filtering, naesseth2017variational}, which is asymptotically unbiased estimator on log-likelihood
\begin{equation}
\text{ELBO}_{\text{SMC}}=\mathbb{E}\bigg[\sum_{t=1}^T\log\bigg(\frac{1}{K}\sum_{k=1}^K\omega_t^k\bigg)\bigg] \label{elbosmc}
\end{equation}
where $K$ is the number of particles and $\omega_t^k$ is the weight of particle $k$ at time $t$. Each particle is defined by weight $\omega_t^k$ and value $\bm{x}_t^p$. At time $t=0$ each particle value $\bm{x}_t^p$ is sampled from initial latent distribution $p_{\theta}(\bm{x}_0)$. In this paper, the transition (emission) function is denoted as $f(\cdot|\cdot)$ ( $g(\cdot|\cdot)$ ). By resampling from the previous particle set $\{\bm{x}_{t-1}^k\}_{k=1}^K$, the weight $\omega_t^k$ for every particle at each time is defined as below
\begin{equation}
\omega_t^k=\frac{f_{\theta}(\bm{x}_t^k|\bm{x}_{t-1}^{\alpha_t^k}, \bm{u}_t, \bm{y}_{t-1})g_{\theta}(\bm{y}_t|\bm{x}_t^k)}{q_{\phi}(\bm{x}_t^k|\bm{x}_{t-1}^{\alpha_t^k}, \bm{y}_t, \bm{u}_t)} \label{particle}
\end{equation}
where $\bm{x}_t^k$ is sampled from the proposal $q_{\phi}(\cdot|\bm{x}_{t-1}^{\alpha_t^k}, \bm{y}_t, \bm{u}_t)$, and the index follows 
\begin{equation}
\alpha_t^k\sim\text{Discrete}\bigg(\bigg\{\frac{\omega_{t-1}^k}{\sum_{l=1}^K\omega_{t-1}^l}\bigg\}_{k=1}^K\bigg) \nonumber
\end{equation} 
with ancestor index $\alpha_1^k=k$. Following \citep{le2017auto} we know that
\begin{equation}
\mathbb{E}\bigg[\prod_{t=1}^T\frac{1}{K}\sum_{k=1}^K\omega_t^k\bigg]:=\mathbb{E}\bigg[\hat{Z}_{\text{SMC}}(\bm{x}_{1:T}^{1:K},\alpha_{1:T-1}^{1:K})\bigg]=p_{\theta}(\bm{y}_{1:T}) \label{zsmc}
\end{equation}
showing that product of particle weights is an unbiased estimator to the marginal likelihood of observations.

\subsection{Hamiltonian Monte Carlo}
In learning transition and emission of SSM, it is key to generate samples from posterior distribution of latent states given observation data. However, the real posterior is intractable in nonlinear SSM. In this work, different previous methods \citep{le2017auto, maddison2017filtering, naesseth2017variational}, in each time step we don't learn proposals but use HMC \citep{neal2011mcmc} to directly sample latent states from the posterior, which can reduce the model complexity significantly. At each time step $t$, we denote the joint log likelihood of the observations and latent states as
\begin{equation}
\mathcal{L}(\bm{x}_t)=\log g(\bm{y}_t|\bm{x}_t) + \log f(\bm{x}_t|\bm{x}_{t-1}, \bm{y}_{t-1}, \bm{u}_t)\label{ll}
\end{equation}
summing log probability of emission and transition function together. Then we add to it a term involving "momenta" variables $\bm{p}_t$, to obtain the Hamiltonian energy function,
\begin{equation}
H(\bm{x}_t, \bm{p}_t)=-\mathcal{L}(\bm{x}_t) + \frac{1}{2}\bm{p}_t^T M^{-1}\bm{p}_t \label{hamilton}
\end{equation}
This quantity can be interpreted as in physical terminology as the sum of the potential energy $\mathcal{L}(\bm{x}_t)$ and the kinetic energy $\frac{1}{2}\bm{p}_t^TM^{-1}\bm{p}_t$, where $M$ acts as the canonical mass matrix. The joint distribution of $\bm{x}_t$ and $\bm{p}_t$ is then defined as $p(\bm{x}_t, \bm{p}_t)\propto \exp(-H(\bm{x}_t, \bm{p}_t))$.

Denote the time derivatives with the dot notation, i.e., $\dot{\bm{x}_t}=d\bm{x}_t/d\tau$, where $\tau$ is the refined time in $[t, t+1)$. The Hamiltonian equations of motion governing the dynamics of this system can be written as
\begin{equation}
\dot{\bm{x}_t}=\frac{\partial}{\partial\bm{p}_t}H(\bm{x}_t, \bm{p}_t), \hspace{20pt} \dot{\bm{p}_t}=-\frac{\partial}{\partial\bm{x}_t}H(\bm{x}_t, \bm{p}_t) \nonumber
\end{equation}
Obviously these equations are time-reversible, and the dynamics conserve the total energy. These continuous-time equations can be discretized to give "leapfrog" algorithms which are used for Monte Carlo simulations along with Metropolis-Hastings correction steps \citep{neal2011mcmc}.

\section{Model Learning Method}
In this section, the proposed Hamiltonian Sequential Monte Carlo (HSMC) is thoroughly described. We first extend Riemann manifold HMC to recurrent setting, and then formulate the learning objective function based on variational inference. Finally, based on HSMC, we will introduce a new meta-learning method for nonstationary state space model. 

\subsection{Riemann Manifold HMC}
The sampling quality of HMC is heavily influenced by the choice of mass matrix \citep{girolami2011riemann}. In order to automatically select mass matrix across different time steps, we adopt Riemann manifold HMC \citep{girolami2011riemann} to incorporate local geometric properties of latent states. Here, at time step $t$, we parametrize the mass matrix as a function of latent states, i.e., $\bm{M}(\bm{x}_t)$, and it is a positive definite metric tensor defining the Riemann manifold on which we are sampling. Defining the kinetic energy in terms of the mass matrix, we can get the Hamiltonian energy function as below
\begin{equation}
H_{\text{rm}}(\bm{x}_t, \bm{p}_t) = -\mathcal{L}(\bm{x}_t)+\frac{1}{2}\bm{p}_t^T\bm{M}(\bm{x}_t)^{-1}\bm{p}_t+\frac{1}{2}\log\{(2\pi)^{D_x}|\bm{M}(\bm{p}_t)|\} \label{hrm}
\end{equation}
where $D_x$ is the dimension of latent state. The desired marginal density of $\bm{x}_t$ can be obtained by integrating out the momenta $\bm{p}_t$. The equations of motion at every time step $t$ are as below
\begin{eqnarray}
\dot{\bm{x}_t}&=&\bm{M}(\bm{x}_t)^{-1}\bm{p}_t \label{dx} \\
\dot{\bm{p}_t}&=&\nabla\mathcal{L}(\bm{x}_t)-\frac{1}{2}\text{tr}\big(\bm{M}(\bm{x}_t)^{-1}\nabla \bm{M}(\bm{x}_t)\big) \nonumber \\
&& +\frac{1}{2}\bm{p}_t^T\bm{M}(\bm{x}_t)^{-1}\nabla \bm{M}(\bm{x}_t)\bm{M}(\bm{x}_t)^{-1}\bm{p}_t\label{dp}
\end{eqnarray}
where the last equation is denoted as a function $\hat{U}(\cdot)$ of latent state. To discretize this system of equations, we use the generalized leapfrog algorithm, where a first order symplectic integrator is composed with its adjoint; the resultant second order integrator can be shown to be both time-reversible and symplectic \citep{leimkuhler2004simulating}. The Riemann manifold HMC operation is formulated as a function shown as below. 
\begin{algorithm}
\caption{Riemann Manifold HMC function: RMHMC($\bm{x}, \bm{p}, \bm{M}, \mathcal{L}, S, \epsilon$;$\phi$)}
\begin{algorithmic}
\REQUIRE{$\bm{M}(\cdot)$: mass matrix function parameterized by $\phi$; $\mathcal{L}(\cdot)$: joint log likelihood function}
\REQUIRE{$S$: number of HMC steps; $\epsilon$: step size ($S\epsilon < 1$)}
\STATE{Define $\hat{U}(\bm{x}, \bm{p})=\nabla\mathcal{L}(\bm{x})-\frac{1}{2}\text{tr}\big(\bm{M}(\bm{x})^{-1}\nabla \bm{M}(\bm{x})\big)+\frac{1}{2}\bm{p}^T\bm{M}(\bm{x})^{-1}\nabla \bm{M}(\bm{x})\bm{M}(\bm{x})^{-1}\bm{p}$}
\FOR{$s\leftarrow 1$ to $S$}
\STATE{$\tilde{\bm{p}}\leftarrow\bm{p}-\epsilon/2\odot\hat{U}(\bm{x}, \bm{p})$}
\STATE{$\bm{x}\leftarrow\bm{x}+\epsilon\odot(\bm{M}(\bm{x})^{-1}\tilde{\bm{p}})$}
\STATE{$\bm{p}\leftarrow\tilde{\bm{p}}-\epsilon/2\odot\hat{U}(\bm{x}, \tilde{\bm{p}})$}
\ENDFOR
\RETURN{augmented latent state $\bm{x}$ and momenta variables $\bm{p}$}
\end{algorithmic}
\end{algorithm}
In implementation, the mass matrix is realized by a low-rank matrix, i.e., $\bm{M}(\bm{x})=\bm{u}(\bm{x})+\bm{\nu}(\bm{x})\bm{\nu}(\bm{x})^T$, where $\bm{u}(\bm{x}),\bm{\nu}(\bm{x})\in\mathbb{R}^{D_x\times 1}$ are MLPs with current latent states as inputs, and their parameters are denoted as $\phi$. Then we denote mass matrix as $\bm{M}_{\phi}$.

\subsection{Hamiltonian SMC}
Here we describe the proposed method, Hamiltonian Sequential Monte Carlo (HSMC). At each time step, the particles of latent states are directly sampled from the transition function conditioned on previous particles $\tilde{\bm{x}}_{t-1}^{\alpha_{t-1}^k}$, current input variables $\bm{u}_t$ and previous observations $\bm{y}_{t-1}$, which plays the role of proposal in previous work \citep{le2017auto, maddison2017filtering, naesseth2017variational}. The momenta variables are introduced necessarily, sampled from Gaussian distribution $\mathcal{N}(0, \bm{M}_{\phi})$. And each particle consists of a tuple of latent state and momenta variable. Then we use the Riemann Manifold Hamiltonian Monte Carlo (RMHMC) to transform the sampled particles to follow the posterior of latent states given current observations. For each particle $k$, denote $(\bm{x}_t^{0,k}, \bm{p}_t^{0,k})$ and $(\bm{x}_t^{S,k}, \bm{p}_t^{S,k})$ as initial particle and transformed particle with $S$-step RMHMC. Due to the measure preserving property of HMC \citep{neal2011mcmc}, the initial and transformed particles have the same density value, i.e., $q^0_t(\bm{x}_t^{0,k}, \bm{p}_t^{0,k})=q^S_t(\bm{x}_t^{S,k}, \bm{p}_t^{S,k})$, even though they follow different distributions. Then the weight for $k$-th particle can be defined as 
\begin{eqnarray}
\omega_t^k&=&\frac{g_{\theta}(\bm{y}_t|\bm{x}_t^{S,k})f_{\theta}(\bm{x}^{S,k}_t|\tilde{\bm{x}}_t^{\alpha^k_{t-1}}, \bm{u}_t, \bm{y}_{t-1})\mathcal{N}(\bm{p}_t^{S,k}|0,\bm{M}_{\phi}(\bm{x}_t^{S,k}))}{q^S_t(\bm{x}_t^{S,k}, \bm{p}_t^{S,k})} \nonumber \\
&=&\frac{g_{\theta}(\bm{y}_t|\bm{x}^{S,k}_t)f_{\theta}(\bm{x}^{S,k}_t|\tilde{\bm{x}}_t^{\alpha^k_{t-1}}, \bm{u}_t, \bm{y}_{t-1})\mathcal{N}(\bm{p}^{S,k}_t|0, \bm{M}_{\phi}(\bm{x}_t^{S,k}))}{f_{\theta}(\bm{x}^{0,k}_t|\tilde{\bm{x}}_t^{\alpha^k_{t-1}}, \bm{u}_t, \bm{y}_{t-1})\mathcal{N}(\bm{p}^{0,k}_t|0, \bm{M}(\bm{x}_t^{0,k}))} \label{weights}
\end{eqnarray}
where the second equation comes from the measure preserving property of HMC. Therefore, the proposed algorithm is summarized as below.
\begin{algorithm}[H]
\caption{Hamiltonian Sequential Monte Carlo (HSMC)}
\label{alg2}
\begin{algorithmic}
\REQUIRE observation data $\bm{y}_{1:T}$, control variables $\bm{u}_{1:T}$ 
\REQUIRE model parameters $\theta$, mass matrix $\mathcal{M}_{\phi}$ with parameters $\phi$.
\STATE{Sample initial particle values $\bm{x}_1^{0,k}\sim p_{1,\theta}(\cdot|\bm{u}_1)$.}
\STATE{Compute particle weights $\omega_1^k={g_{\theta}(\bm{y}_1|\bm{x}_1^{0,k})}$.}
\STATE{Initialize particle set $\tilde{\bm{x}}_1^k\leftarrow\bm{x}_1^{0,k}$}
\FOR{$t=2,\ldots,T$}
\STATE{Sample ancestor index $\alpha^k_{t-1}\sim\text{Discrete}(\cdot|\omega_{t-1}^1,\ldots,\omega_{t-1}^K)$.}
\STATE{Sample particle value from transition function $\bm{x}_t^{0,k}\sim f_{\theta}(\cdot|\bm{u}_t, \bm{y}_{t-1}, \tilde{\bm{x}}_{t-1}^{\alpha_{t-1}^k})$.}
\STATE{Sample momenta variables from Gaussian distribution $\bm{p}_t^k\sim\mathcal{N}(0, \bm{M}_{\phi}(\bm{x}_t^{0,k}))$.}
\STATE{Process each particle value by RMHMC operation 
\begin{equation}
\bm{x}_t^{S,k}, \bm{p}_t^{S,k}=\text{RMHMC}(\bm{x}_t^{0,k}, \bm{p}^{0,k}_t, \bm{M}_{\phi}, \mathcal{L}, S, \epsilon;\phi) \label{rmhmc}
\end{equation}}
\STATE{Update particle set $\tilde{\bm{x}}_t^k\leftarrow\{\tilde{\bm{x}}^k_t, \bm{x}_t^{S,k}\}$} 
\STATE{Compute particle weights $\omega_t^k$ in \eqref{weights}.}
\ENDFOR
\STATE{Compute marginal likelihood $\hat{Z}_{\text{HSMC}}=\prod_{t=1}^T\frac{1}{K}\sum_{k=1}^K\omega_t^k$}
\RETURN{particles $\bm{x}_{1:T}^{1:K}$, weights $\omega_{1:T}^{1:K}$ and likelihood estimate $\hat{Z}_{\text{HMSC}}$.}
\end{algorithmic}
\end{algorithm}

\subsection{Objective Function}
In this work, we choose the objective function to be the ELBO defined as SMC marginal likelihood estimator in \eqref{elbosmc}. However, we can show that incorporation of HMC can make the ELBO objective arbitrarily tight. Defining $\hat{Z}_{\text{HSMC}}:=\prod_{t=1}^T\frac{1}{K}\sum_{k=1}^K\omega_t^k$, the ELBO objective can be formulated as 
\begin{eqnarray}
\lefteqn{\text{ELBO}_{\text{HSMC}}(\theta, \phi, S, K, \epsilon, \bm{u}_{1:T}, \bm{y}_{1:T})} \nonumber \\
&&=\int Q_{\text{HSMC}}(\bm{x}^{S,1:K}_{1:T}, \bm{p}^{S,1:K}_{2:T}, \alpha_{1:T-1}^{1:K};\bm{u}_{1:T}, \bm{y}_{1:T})\bigg(\log\hat{Z}_{\text{HSMC}}\bigg) d\bm{x}_{1:T}^{S,1:K}d\bm{p}^{S,1:K}_{2:T}d\alpha_{1:T-1}^{1:K} \label{hsmcelbo}
\end{eqnarray}
where $Q_{\text{HSMC}}$ is formulated as 
\begin{eqnarray}
\lefteqn{Q_{\text{HSMC}}(\bm{x}_{S,1:T}^{1:K}, \bm{p}_{2:T}^{S,1:K}, \alpha_{1:T-1}^{1:K};\bm{u}_{1:T}, \bm{y}_{1:T})} \nonumber \\
&&= \bigg(\prod_{k=1}^K p_{1}(\bm{x}_1^{S,k}|\bm{u}_1)\bigg)\bigg(\prod_{t=2}^T\prod_{k=1}^K q^S_{t,\phi}(\bm{x}_t^{S,k}, \bm{p}_t^{S,k}\big| \bm{u}_t,\bm{y}_{t-1}, \tilde{\bm{x}}_{t-1}^{\alpha^k_{t-1}})\cdot\text{Discrete}(\alpha_{t-1}^k\big|\omega_{t-1}^{1:K})\bigg)\nonumber
\end{eqnarray}
where $\bm{x}_t^{S,k}, \bm{p}_t^{S,k}$ are obtained as \eqref{rmhmc} and $q^S_{t,\phi}$ is the distribution of particle and momenta variables after Hamiltonian dynamics. 

We adopt stochastic gradient descent \citep{hoffman2013stochastic} to learn optimal model and mass matrix parameters $\theta, \phi$. The gradient of objective function can be derived as below, where observed and control data $\bm{y}_{1:T}, \bm{u}_{1:T}$ are omitted here,
\begin{eqnarray}
\lefteqn{\nabla_{\theta,\phi}\text{ELBO}_{\text{HSMC}}} \nonumber \\
&=&\nabla_{\theta,\phi}\int Q_{\text{HSMC}}(\bm{x}^{S,1:K}_{1:T}, \bm{p}^{S,1:K}_{2:T}, \alpha_{1:T-1}^{1:K})\bigg(\log\hat{Z}_{\text{HSMC}}\bigg) d\bm{x}_{1:T}^{S,1:K}d\bm{p}^{S,1:K}_{2:T}d\alpha_{1:T-1}^{1:K} \nonumber \\
&=&\int\nabla_{\theta,\phi}Q_{\text{HSMC}}(\bm{x}^{S,1:K}_{1:T}, \bm{p}^{S,1:K}_{2:T}, \alpha_{1:T-1}^{1:K})\bigg(\log\hat{Z}_{\text{HSMC}}\bigg) \nonumber \\
&&+Q_{\text{HSMC}}(\bm{x}^{S,1:K}_{1:T}, \bm{p}^{S,1:K}_{2:T}, \alpha_{1:T-1}^{1:K})\nabla_{\theta,\phi}\log\hat{Z}_{\text{HSMC}}d\bm{x}_{1:T}^{S,1:K}d\bm{p}^{S,1:K}_{2:T}d\alpha_{1:T-1}^{1:K} \nonumber
\end{eqnarray}
\begin{eqnarray}
&=&\int Q_{\text{HSMC}}(\bm{x}^{S,1:K}_{1:T}, \bm{p}^{S,1:K}_{2:T}, \alpha_{1:T-1}^{1:K})\bigg[\nabla_{\theta,\phi}\log Q_{\text{HSMC}}(\bm{x}^{S,1:K}_{1:T}, \bm{p}^{S,1:K}_{2:T}, \alpha_{1:T-1}^{1:K})\log\hat{Z}_{\text{HSMC}} \nonumber \\
&&+\nabla_{\theta,\phi}\log\hat{Z}_{\text{HSMC}} \bigg] d\bm{x}^{S,1:K}_{2:T}d\bm{p}^{S,1:K}_{1:T}d\alpha_{1:T-1}^{1:K} \nonumber 
\end{eqnarray}
In order to reduce the gradient variance, we ignore the first term in the squared bracket above. 

\subsection{Theoretical Analysis}
Based on the property of importance sampling \citep{murphy2012machine}, we can easily show that at each time step $t$ the weight expression \eqref{weights} is an unbiased estimator of marginal likelihood likelihood of observations $\bm{y}_t$ conditioned on $\bm{u}_t$ and $\bm{y}_{t-1}$. In this section, we show that as step number $S$ increasing with particle number $K$ fixed, our learning objective \eqref{hsmcelbo} can be arbitrarily close or converge to the marginal log likelihood of observations. As \citep{le2017auto,maddison2017filtering} we can assume the state space model has independent structure $p_{\theta}(\bm{x}_{1:t-1}|\bm{y}_{1:t})=p_{\theta}(\bm{x}_{1:t-1}|\bm{y}_{1:t-1})$ for $t=2,\ldots,T$. It is a reasonable assumption since in online learning and many practical applications, the future observations can not be known in advance. Define conditional marginal likelihood of observations for each time $t$ and particle $k$ as
\begin{equation}
\int f_{\theta}(\bm{x}|\tilde{\bm{x}}_{t-1}^{\alpha_t^k}, \bm{u}_t, \bm{y}_{t-1})g_{\theta}(\bm{y}_t|\bm{x})\mathcal{N}(\bm{p}|0, M(\bm{x})) d\bm{x}d\bm{p} := p_{\theta}(\bm{y}_t|\tilde{\bm{x}}_{t-1}^{\alpha_t^k}, \bm{u}_t, \bm{y}_{t-1}) \nonumber
\end{equation}
Then, if we can show the convergence of ELBO at each time $t$ and particle $k$, the convergence of ELBO objective across all time in \eqref{hsmcelbo} can be shown. 

According to the Hamilton energy defined in \eqref{hamilton}, the HMC adopted in algorithm \ref{alg2} is ergodic with invariant distribution as
\begin{equation}
p_{\theta}(\bm{x}, \bm{p}\big|\bm{u}_t,\bm{y}_{t-1}, \tilde{\bm{x}}_{t-1}^{\alpha^k_{t-1}}) \propto f_{\theta}(\bm{x}|\tilde{\bm{x}}_{t-1}^{\alpha_t^k}, \bm{u}_t, \bm{y}_{t-1})g_{\theta}(\bm{y}_t|\bm{x})\mathcal{N}(\bm{p}|0, M(\bm{x}))\label{statdist}
\end{equation}
which is the posterior of joint probabilities of emission, transition and momenta distributions, or alternatively,
\begin{equation}
p_{\theta}(\bm{x}, \bm{p}\big|\bm{u}_t,\bm{y}_{t-1}, \tilde{\bm{x}}_{t-1}^{\alpha^k_{t-1}}) = \frac{f_{\theta}(\bm{x}|\tilde{\bm{x}}_{t-1}^{\alpha_t^k}, \bm{u}_t, \bm{y}_{t-1})g_{\theta}(\bm{y}_t|\bm{x})\mathcal{N}(\bm{p}|0, M(\bm{x}))}{p_{\theta}(\bm{y}_t|\tilde{\bm{x}}_t^{\alpha_{t-1}^k}, \bm{u}_t, \bm{y}_{t-1})}\label{statdist2}
\end{equation}
Based on properties of HMC \citep{neal2011mcmc}, the particle distribution will tend to invariant distribution in total variation, with the increase of step number,i.e., for all $t$ and $k$,
\begin{equation}
\lim_{S\to\infty}\|q^S_{t,\phi}(\bm{x}, \bm{p})-p_{\theta}(\bm{x}, \bm{p}|\bm{u}_t,\bm{y}_{t-1}, \tilde{\bm{x}}_{t-1}^{\alpha^k_{t-1}})\|_{\text{TV}}=0 \label{tv}
\end{equation}
Then based on the weight definition \eqref{weights}, we have, for each $t$ and $k$,
\begin{eqnarray}
\lefteqn{\lim_{S\to\infty}\int q^S_{t,\phi}(\bm{x}, \bm{p})\log\big(\omega^k_t\big) d\bm{x} d\bm{p}} \nonumber \\
&=&\int q^{\infty}_{t,\phi}(\bm{x}, \bm{p})\log\bigg(\frac{f_{\theta}(\bm{x}|\tilde{\bm{x}}_{t-1}^{\alpha_t^k}, \bm{u}_t, \bm{y}_{t-1})g_{\theta}(\bm{y}_t|\bm{x})\mathcal{N}(\bm{p}|0, M(\bm{x}))}{q^{\infty}_{t,\phi}(\bm{x}, \bm{p})}\bigg) d\bm{x} d\bm{p} \nonumber \\
&=&\log p_{\theta}(\bm{y}_t|\tilde{\bm{x}}_t^{\alpha_{t-1}^k}, \bm{u}_t, \bm{y}_{t-1}) \label{inftyupper}
\end{eqnarray}
where the second equality is due to \eqref{statdist2} and \eqref{tv}. Due to the concavity of log function, the ELBO objective \eqref{hsmcelbo} at each time $t$ can be lower bounded as
\begin{eqnarray}
\lefteqn{\int \prod_{k=1}^K q^{S}_{t,\phi}(\bm{x}^{S,k}_t,\bm{p}_t^{S,k})\log\bigg[\frac{1}{K}\sum_{k=1}^k\omega_t^k\bigg]d\bm{x}_t^{S,1:K}d\bm{p}_t^{S,1:K}} \nonumber \\
&\ge&\int \prod_{k=1}^K q^{S}_{t,\phi}(\bm{x}^{S,k}_t,\bm{p}_t^{S,k})\bigg[\frac{1}{K}\sum_{k=1}^K\log\omega^k_t\bigg] d\bm{x}_t^{S,1:K}d\bm{p}_t^{S,1:K} \nonumber \\
&=&\int q^S_{t,\phi}(\bm{x}, \bm{p})\log(\omega^k_t)d\bm{x}d\bm{p} \label{lowerb}
\end{eqnarray}
Combining \eqref{inftyupper} and \eqref{lowerb} yields that at each time $t$ the ELBO objective converges to conditional log likelihood of observations. Due to the independent structure of latent state space, we can show that the overall objective \eqref{hsmcelbo} converges to the log likelihood of observations $p(\bm{y}_{1:T})$ with the increase of step number $S$. The empirical study tells us that the performance is good enough when $S$ is just around 10. 

\section{Improving Gaussian Process State Space Model}
Gaussian Process State Space Models (GP-SSM) are a popular class of stochastic SSMs \citep{frigola2013bayesian, frigola2014variational, eleftheriadis2017identification, doerr2018probabilistic, ialongo2018non}. In SSM, at each time step the system is taken to evolve as a Markov chain by the transition function, mapping a latent state to the next. By placing a Gaussian Process (GP) prior on the transition function, we can obtain the Gaussian process state-space model (GP-SSM), which is a fully Bayesian non-parametric treatment on modeling problem. It has many advantages: 1) better uncertainty estimates based on the posterior of the transition function; 2) avoiding overfitting with little amount of data; 3) handling large amount of data without model saturation.

\subsection{Preliminary}
Same as standard SSM, we model the sequence of observations $\bm{Y}=\{\bm{y}_t\}_{t=1}^T$ by a corresponding sequence of latent states $\bm{X}=\{\bm{x}_t\}_{t=1}^T$, and $\bm{x}_t\in\mathbb{R}^{D_x}, \bm{y}_t\in\mathbb{R}^{D_y}$. Here the state transition is assumed to be governed by a nonparametric stochastic function $\bm{f}\in\mathbb{R}^{D_x}$ following a GP prior. Specifically, we have
\begin{equation}
\bm{x}_{t+1}\sim\mathcal{N}(\bm{f}(\bm{x}_t),\bm{Q}), \hspace{15pt} f_d\sim\mathcal{G}\mathcal{P}(0, k_d(\cdot,\cdot)), \hspace{15pt} d=1,\ldots, D_x \nonumber
\end{equation}
where $\bm{Q}$ is the variance matrix and every $d$-th latent dimension has its own GP function $f_d$. The emission function is still parametric same as \eqref{model}. In order to reduce the computation complexity in learning GP, we adopt the induced-inputs method \citep{snelson2006sparse} and variational sparse GP \citep{titsias2009variational} in model learning, which achieves success in many practical problems. For every GP function $f_d$, we first introduce $P$ inducing GP targets $\bm{z}_d=[z^1_d,\ldots,z^P_d]$ at inducing GP inputs $\zeta_d=[\zeta^1_d,\ldots,\zeta^P_d]$, which are jointly Gaussian with the transition function $f_d$. Then for every latent dimension, the true GP predictive distribution can be approximated by using the set inducing inputs and outputs as below, with notation $d$ omitted,
\begin{equation}
p(f^*_d|\bm{x}^*,\bm{f},\bm{X})\approx p(f^*|\bm{x}^*,\bm{z},\bm{\zeta}), \hspace{15pt} p(\bm{z})=\mathcal{N}(\bm{z}|0, \bm{K}_{\zeta, \zeta}) \label{gpprior}
\end{equation}
where the covariance matrix $\bm{K}_{\zeta, \zeta}$ with entries $K_{ij}=k(\zeta_i,\zeta_j)$. Following \eqref{model} we show the joint distribution of GP-SSM for completeness,
\begin{equation}
p(\bm{y}_{1:T},\bm{x}_{1:T},\bm{f}_{2:T},\bm{z})=\bigg[\prod_{t=1}^Tg_{\theta}(\bm{y}_t|\bm{x}_t)\bigg]p(\bm{x}_1)p(\bm{z})\bigg[\prod_{t=2}^Tp(\bm{x}_t|\bm{f}_t)p(\bm{f}_t|\bm{x}_{t-1}, \bm{u}_t, \bm{y}_{t-1},\bm{z})\bigg] \label{gpjoint}
\end{equation}
where $p(\bm{f}_t|\hat{\bm{x}}_{t-1},\bm{z})=\prod_{d=1}^{D_x}p(f_{t,d}|\bm{x}_{t-1}, \bm{u}_t, \bm{y}_{t-1}, \bm{z}_d)$ and $\bm{z}:=[\bm{z}_1,\ldots,\bm{z}_{D_x}]$. Here $p(\bm{x}_t|\bm{f}_t)=\mathcal{N}(\bm{x}_t|\bm{f}_t, \text{diag}(\sigma^2_{x,1}, \ldots, \sigma^2_{x,D_x})$ and $p(\bm{x}_1), p(\bm{z})$ are assumed to be Gaussian.

\subsection{Motivation}

As far as we know, all previous work on GP-SSM \citep{frigola2013bayesian, frigola2014variational, frigola2015bayesian, eleftheriadis2017identification, doerr2018probabilistic, ialongo2018non} assume the emission function to be a linear mapping between latent state $\bm{x}_t$ and the mean of observation $\bm{y}_t|\bm{x}_t$. However, in many practical applications the emission involves nonlinear function of $\bm{x}_t$ \citep{gultekin2017nonlinear}. In this case the posterior of latent states is non-Gaussian and possibly multi-modal. So, the Gaussian distribution of transition function cannot approximate the real posterior well enough. 

In previous work \citep{frigola2013bayesian, frigola2014variational, ialongo2018non} the variational distributions need to optimize parameter matrices $\bm{A}_t, \bm{b}_t$ and $\bm{S}_t$, which increases model and learning complexity. Although authors in  \citep{doerr2018probabilistic} directly use transition prior as variational approximation for latent posterior without extra parameters, it cannot exploit information contained in the current observations other than by adapting $q(\bm{u})$, which cannot handle cases with high observation noise and long sequence length \citep{ialongo2018non}.

\subsection{Variational Inference for GP-SSM with HSMC}
Here we use HSMC to solve problems mentioned above. We first design the variational distribution. Following the structure of real latent posterior, the latent states and transition function are not factorized, i.e., $q(\bm{X}, f) = q(\bm{X}|f)q(f)$. By introducing induced outputs $\bm{z}$, the variational distribution can be defined as
\begin{equation}
q(\bm{x}_{1:T},\bm{f}_{2:T},\bm{z})=q(\bm{x}_1)\bigg[\prod_{t=2}^Tp(\bm{x}_t|\bm{f}_t)\prod_{d=1}^{D_x}p(f_{x,d}|\hat{\bm{x}}_{t-1}, \bm{z}_d)\bigg]\cdot\bigg[\prod_{d=1}^{D_x}q(\bm{z}_d)\bigg] \label{gpvar}
\end{equation}
where $\hat{\bm{x}}_{t-1}:=(\bm{x}_{t-1},\bm{u}_t,\bm{y}_{t-1})$, and $q(\bm{x}_1)$ and $q(\bm{z}_d)$ are assumed to be Gaussian, i.e., $q(\bm{z}_d)=\mathcal{N}(\bm{\mu}_d, \bm{\Sigma}_d)$ where $\bm{\mu}_d\in\mathbb{R}^P$ and $\bm{\Sigma}_d\in\mathbb{R}^{P\times P}$. We can further simplify the variational inference by integrating out induced-outputs $\bm{z}$ \citep{hensman2013gaussian}. Then following \eqref{gpjoint} the variational transition function can be derived as
\begin{equation}
q(\bm{x}_t|\hat{\bm{x}}_{t-1})=\int \bigg[p(\bm{x}_t|\bm{f}_t)p(\bm{f}_t|\hat{\bm{x}}_{t-1}, \bm{z})\prod_{d=1}^{D_x}q(\bm{z}_d)\bigg] d\bm{f}_td\bm{z}=\mathcal{N}(\tilde{\bm{\mu}}, \tilde{{\Sigma}}) \nonumber
\end{equation}
where $\tilde{\bm{\mu}}=[\tilde{\mu}_1,\ldots,\tilde{\mu}_{D_x}]$ and $\tilde{{\Sigma}}=\text{diag}(\tilde{\sigma}_1^2,\ldots,\tilde{\sigma}_{D_x}^2)$, for $d=1,\ldots,D_x$,
\begin{eqnarray}
\tilde{\mu}_d&=&k_{\hat{\bm{x}}_{t-1},\bm{\zeta}_d}K^{-1}_{\bm{\zeta}_d,\bm{\zeta}_d}\bm{\mu}_d \nonumber \\
\tilde{\sigma}^2_d &=& k_{\hat{\bm{x}}_{t-1}, \hat{\bm{x}}_{t-1}}-k_{\hat{\bm{x}}_{t-1}, \bm{\zeta}_d}K^{-1}_{\bm{\zeta}_d, \bm{\zeta}_d}(K_{\bm{\zeta}_d, \bm{\zeta}_d}-\Sigma_d)K^{-1}_{\bm{\zeta}_d, \bm{\zeta}_d}k_{\hat{\bm{x}}_{t-1}, \bm{\zeta}_d}^T+ \sigma^2_{x,d},\hspace{15pt}  \nonumber 
\end{eqnarray}
In this work, we use $K$-particle HSMC to approximate the intractable posterior. Denote $\hat{\bm{x}}_{t-1}^k=(\bm{x}_{t-1}^{\alpha^k_{t-1}}, \bm{u}_t, \bm{y}_{t-1})$ where $\alpha_{t-1}^k$ is the re-sampling index. At each time step $t$, we first sample multiple particles of latent states $\bm{x}_t^{0,k}$ and momenta $\bm{p}_t^{0,k}$ from the transition function $q(\bm{x}_t|\hat{\bm{x}}^k_{t-1})$ and Gaussian $\mathcal{N}(0,\bm{M}(\bm{x}_t^{0,k}))$ respectively. Then by using $S$-step RMHMC operation we transform the sampled particles to approximate the posterior as 
\begin{equation}
p_{\theta}(\bm{x},\bm{p}|\hat{\bm{x}}_{t-1})\propto g_{\theta}(\bm{y}_t|\bm{x})q(\bm{x}|\hat{\bm{x}}_{t-1})\mathcal{N}(\bm{p}|0,\bm{M}(\bm{x})) \nonumber
\end{equation}
Due to the measure preserving and reversibility property of HMC, we define the particle weights as 
\begin{equation}
\omega_t^k=\frac{g_{\theta}(\bm{y}_t|\bm{x}_t^{S,k})q(\bm{x}_t^{S,k}|\hat{\bm{x}}_{t-1}^k)\mathcal{N}(\bm{p}_t^{S,k}|0,\bm{M}(\bm{x}^{S,k}_t))}{q(\bm{x}_t^{0,k}|\hat{\bm{x}}_{t-1}^k)\mathcal{N}(\bm{p}_t^{0,k}|0,\bm{M}(\bm{x}_t^{0,k}))} \label{gpweights}
\end{equation}
According to the joint and variational distribution in \eqref{gpjoint}\eqref{gpvar}, the ELBO can be computed as
\begin{eqnarray}
\text{ELBO}_{\text{GP-SSM}}&=&\sum_{t=1}^T\int\prod_{k=1}^K q(\bm{x}_t^{S,k}|\hat{\bm{x}}_{t-1}^k)\log\bigg(\frac{1}{K}\sum_{k=1}^K\omega_t^k\bigg)d\bm{x}_t^{0,1:K} \nonumber \\
&&-\sum_{d=1}^{D_x}\text{KL}\big(q(\bm{z}_d)\| p(\bm{z}_d))\big) \nonumber
\end{eqnarray}
where $p(\bm{z}_d)$ is defined in \eqref{gpprior}. In experimental study, we show that without increasing model complexity, HSMC can improve learning performance on GP-SSM with nonlinear emission.

\section{Experiments}
In this section, we conduct experiments to show the benefits of proposed algorithms. First, we use synthetic data to verify the advantages of HSMC over conventional variational SMC, where both transition and emission are realized by neural networks. Then based on real-world dataset, we show that HSMC can better learn GP-SSM when emission function is nonlinear. 

\subsection{Synthetic Data}
The synthetic data is generated by a nonlinear state-space model 
\begin{equation}
\bm{x}_{t+1}=\bm{A}\bm{x}_t+\bm{\epsilon}_t, \hspace{20pt} \bm{y}_t=g_{\theta}(\bm{x}_t)+\bm{\xi}_t \nonumber
\end{equation} 
where $\bm{x}_t\in\mathbb{R}^{10}, \bm{y}_t\in\mathbb{R}^{30}$ and $g_{\theta}$ is realized by two-layer neural network with 20 hidden neurons. The non-linearities in $g_{\theta}$ are ReLU and Sigmoid. And the noise $\bm{\epsilon}_t, \bm{\xi}_t$ follow independent Gaussian with variance of $0.2$. Every sequence has length of 100. Both training and testing datasets only contain observations $\bm{y}_t$.

In this experiment, the base models are variational recurrent neural network (VRNN) \citep{chung2015recurrent} and stochastic recurrent neural network (SRNN) \citep{fraccaro2016sequential}, where the emission is realized by deep neural network and transition is implemented by gated recurrent unit (GRU) \citep{chung2014empirical}. The comparison is between the base models (VRNN or SRNN) augmented by FIVO \citep{maddison2017filtering} and HSMC. And generation network in every compared model is the same. However, in base model with HSMC, the proposal neural network is omitted. The performance metric is the log-likelihood per step. Here $K$ is the number of particles and $S$ is the number of step in RMHMC. Performance of FIVO is not related with $S$.

\begin{table}[H]
\centering
\caption{Performance Comparison in Synthetic Data}
\begin{tabular}
{c|c|c|c|c}
          & $K=10, S=5$ & $K=10, S=10$ & $K=5, S=5$ & $K=5, S=10$ \\\hline\hline
VRNN-FIVO &  2.56     &    2.56    &    2.12  &   2.12    \\\hline        
VRNN-HSMC &  2.78     &    2.86    &    2.37  &   2.54    \\\hline        
SRNN-FIVO &  2.64     &    2.64    &    2.31  &   2.31 \\\hline        
SRNN-HSMC &  2.80     &    2.86    &    2.49  &   2.53         
\end{tabular}
\end{table}

\subsection{Bike-sharing Demand Data}
In this experiment, we use bike-sharing record data in New York city \citep{dataset}, from July 2014 to July 2017. In every transaction record, there are trip duration, bike check out/in time, names of start and end stations, and customer information such as age and gender. We first aggregate transaction records to bike demands at each station, ignoring customer information. Then we further remove stations existing for less than two years. And stations with less than one bike used per hour are also deleted. Finally, there are only 269 stations left in the dataset.  

Here we compare the orginal GP-SSM \citep{doerr2018probabilistic} with multiple particles (GP-SSM-SMC) and that augmented by HSMC (GP-SSM-HSMC). Both models have same latent dimension and number of inducing points. The performance metric is log likelihood. The results are shown below. We also the results of VRNN-FIVO for comparison.

\begin{table}[H]
\centering
\caption{Performance Comparison in Bike-sharing Demand Data}
\begin{tabular}
{c|c|c|c|c}
          & $K=10, S=5$ & $K=10, S=10$ & $K=5, S=5$ & $K=5, S=10$ \\\hline\hline
GP-SSM-SMC &  -653.2     &    -653.2    &    -668.7  &   -668.7    \\\hline        
GP-SSM-HSMC &  -607.1     &    -591.3    &    -621.4  &   -610.9     \\\hline
VRNN-FIVO  &   -589.2    &   -589.2  & -599.1 & -599.1
\end{tabular}
\end{table}

We find that HSMC can help diminish the performance gap between GP model and neural network model, even though GP has much less parameters than neural networks.

\bibliographystyle{apalike}
\bibliography{hsmc}

\end{document}